# Transcending Classical Neural Network Boundaries: A Quantum-Classical Synergistic Paradigm for Seismic Data Processing

Authors — Zhengyi Yuan, Xintong Dong∗ , Xinyang Wang, Zheng Cong, Shiqi Dong.

College of Instrumentation and Electrical Engineering, Jilin University, Changchun 130026, China.

Corresponding author: Xintong Dong, 18186829038@163.com

ABSTRACT

In recent years, a number of neural-network (NN) methods have exhibited good performance in seismic data processing, such as denoising, interpolation, and frequency-band extension. However, these methods rely on stacked perceptrons and standard activation functions, which imposes a bottleneck on the representational capacity of deep-learning models, making it difficult to capture the complex and non-stationary dynamics of seismic wavefields. Different from the classical perceptron-stacked NNs which are fundamentally confined to real-valued Euclidean spaces, the quantum NNs leverage the exponential state space of quantum mechanics to map the features into high-dimensional Hilbert spaces, transcending the representational boundary of classical NNs. Based on this insight, we propose a quantum-classical synergistic generative adversarial network (QC-GAN) for seismic data processing, serving as the first application of quantum NNs in seismic exploration. In QC-GAN, a quantum pathway is used to exploit the high-order feature correlations, while the convolutional pathway specializes in extracting the waveform structures of seismic wavefields. Furthermore, we design a QC feature complementarity loss to enforce the feature orthogonality in the proposed QC-GAN. This novel loss function can ensure that the two pathways encode non-overlapping information to enrich the capacity of feature representation. On the whole, by synergistically integrating the quantum and convolutional pathways, the proposed QC-GAN breaks the representational bottleneck inherent in classical GAN. Experimental results on denoising and interpolation tasks demonstrate that QC-GAN preserves wavefield continuity and amplitude-phase information under complex noise conditions. To further validate the versatility of QC synergistic paradigm, we extend this paradigm to the UNet architecture, and the QC-UNet shows superior performance in the low-frequency extrapolation compared with the classical U-Net. These cross-task and cross-architecture experiments demonstrate the potential of the QC synergistic paradigm to become a general-purpose technique for seismic data processing.

## INTRODUCTION

Seismic data serve as crucial observational evidence of the physical properties of the subsurface media[1]. High-resolution seismic imaging often requires noise- free and complete data with broad frequency bandwidth. However, numerous field seismic data inevitably suffer from the noise contamination, incomplete geometry due to acquisition constraints, and insufficient low-frequency information limited by the bandwidth of geophones and absorption of subsurface media. Consequently, how to restore the degraded seismic data is a pivotal research direction in seismic exploration. Existing restoration approaches can be broadly categorized into theory-driven and data- driven methods. Theory-driven approaches primarily rely on the physical principles governing the propagation of seismic waves, employing mathematical modeling or incorporating physics-based constraints to reconstruct the degraded seismic data[2]. Specifically, traditional seismic denoising and interpolation techniques include predictive filters[3], decomposition-based methods[4-6], sparse transforms[7], and low-rank-based methods[8]. Basis pursuit inversion[9], deconvolution[10], and envelope inversion[11] serve as representative methods for recovering the missing low-frequency components of seismic data. Although these restoration methods are theoretically capable of delivering the optimal solutions, their performance is often degraded when facing certain complex situations, such as low signal-to-noise ratio, consecutively missing case, and complex seismic wavefield. Data-driven methods rely directly on observational data rather than strict physical assumptions, offering greater flexibility and improved adaptability to more complex cases[12].

In recent years, deep learning (DL) has emerged as a powerful data-driven tool by constructing multilayer perceptrons combined with nonlinear activation functions to emulate the information-processing mechanisms of biological neural systems. These models are capable of automatically extracting patterns from the data itself, so as to perform prediction and decision-making tasks, Consequently, numerous DL-based methods have exhibited good performance in computer vision[13-16] and natural language processing[17-19], which introduces new perspectives and solutions for restoring the degraded seismic data[20-30]. In the early stage, researchers explored the application of convolutional neural network (CNN) to seismic data restoration. For instance, Yu et al.[31] utilized a CNN framework with different datasets to automatically attenuate random noise, linear noise, and multiples, demonstrating the versatility of DL in handling different noise patterns. Dong et al.[32] proposed an optimized denoising CNN framework featuring the construction of adaptive noise dataset, which suppressed the low-frequency noise while preserving the details of valid signals. Sun et al.[33] proposed a CNN-based low-frequency extrapolation method, offering a robust solution to complex conditions arising from random noise, inaccurate source wavelets, and discrepancies in finite-difference schemes. Wang et al.[34] designed an eight-layer residual learning network with optimized back-propagation capabilities to perform anti-aliasing interpolation for regularly missing seismic data. Subsequently, experts have dedicated efforts to further enhancing the performance and generalization of DL-based models, and proposed a number of CNN-based variants for seismic data restoration. Sun et al.[35] incorporated a residual learning strategy within a generative adversarial network (GAN) to achieve high-fidelity denoising of seismic data. Collazos et al.[36] proposed a conditional GAN to perform seismic data interpolation, effectively recovering the missing traces in shot-gather domain. Ovcharenko et al.[37] explored the use of a U-Net architecture to perform frequency extrapolation at the shot-gather level, achieving excellent experimental results on sections from the British Petroleum and SEG Advanced Modeling benchmark models. Dong et al.[22] introduced a non-local attention block and fused it into a multi-scale encoder-decoder architecture, specifically designed to restore the low-components by capturing the long-range spatial dependencies of seismic data.

The advent of self-attention-based transformers has introduced a new perspective to seismic data processing. Compared with CNNs, Transformers can capture more useful global features[38]. Building upon this, Zhang et al.[39] integrated the swin transformer with GAN to develop a novel denoising model. This model leverages the self-attention mechanism to capture the correlations between local and non-local seismic features, thereby enhancing the denoising performance. Wang et al.[40] proposed a hybrid CNN-Transformer network to learn the multi-resolution features of seismic data and applied it to 3D seismic data interpolation. Cong et al.[41] introduced a low-frequency extrapolation transformer and deployed it as an end-to-end DL framework, showing improved extrapolation performance compared with CNN-based architectures. These aforementioned neural networks (NNs) primarily rely on stacking neurons with nonlinear activation functions to increase the representation capacity of models. This highly parameter-dependent architecture inherently restricts their feature extraction and generalization capabilities, encountering bottlenecks when dealing with complex seismic data characterized by high-order nonlinearity and nonstationarity.

The quantum NNs[42] offer a solution to this architectural limitation. By leveraging the quantum mechanical properties such as superposition and entanglement, the quantum NNs map data into a high-dimensional Hilbert space for linear evolution, thereby enabling the efficient modeling of complex data distributions[43]. Compared to classical NNs, quantum NNs exhibit two distinct advantages. First, they possess superior nonlinear representational capacity, allowing them to capture high-dimensional mapping relationships that classical NNs struggle to extract[44]. Second, they require significantly fewer training parameters to achieve comparable or superior accuracy compared with classical NNs, which significantly alleviates the computational burden and mitigates the risk of overfitting[45]. Therefore, quantum NNs have demonstrated significant advantages over classical NNs across certain domains[46-48]. Moreover, recent studies have explored quantum-classical (QC) synergistic architectures, successfully leveraging the complementary strengths of the two paradigms[49]. Motivated by this, we propose a QC-GAN for seismic data processing. To be specific, the QC-GAN architecture employs a dual-pathway design: a quantum pathway dedicated to extracting high-dimensional quantum-state features, operating in conjunction with a classical convolutional pathway that preserves wavefield continuity and seismic morphology. To enforce of these two pathways, we design a QC feature complementarity loss function with orthogonality constraints to ensure these two parallel paths learn distinct yet cooperative representations. By explicitly suppressing feature redundancy, this loss function facilitates the deep fusion of high-dimensional quantum states and classical spatial priors, ultimately empowering the QC-GAN to recover complex seismic events. Experiments on field seismic datasets demonstrate that the proposed QC-GAN, outperforms its classical counterpart in denoising and interpolation tasks, particularly in the weak-signal restoration. Furthermore, to test the generalizability and applicability of this QC synergistic paradigm from both architectural and task-oriented perspectives, we extend this paradigm to the classical UNet and apply the QC-UNet to extrapolate the missing low-frequency components of field ocean bottom cable (OBC) data, demonstrates the potential of QC synergistic paradigm to serve as a highly generalized and robust framework for seismic data processing.

## METHOD

To bridge the gap between quantum computing and geophysics, This section initially review the computational paradigms of quantum NNs. The discussion then transitions to a rigorous exposition of the QC-GAN, elucidating the design of the QC feature complementarity loss function.

**Quantum NNs**

The quantum NNs constructed in this study is designed to efficiently learn complex patterns within high-dimensional Hilbert spaces. Consider an input tensor $X \in \mathbb{R}^{B \times C \times T \times S}$, where $B$, $C$, $T$, and $S$ represent the batch size, number of channels, number of temporal samples, and number of seismic traces, respectively. The input $X$ is first partitioned by performing a sliding-window unfolding along the spatial dimension. The window size $k$ and stride $s$ are configured to match the number of qubits ($N_q$), which is set to 4 in this work. This process generates a sequence of sub-blocks, which are stacked to form the unfolded tensor $X_{unf} \in \mathbb{R}^{M \times N_q}$, where $M = B \times C \times T' \times S'$ represents the total number of patches:

$$X_{unf} = \text{unfold}(X, k, s) \tag{1.1}$$

To process this data, we treat each row of $X_{unf}$ as an individual input vector. $x' \in \mathbb{R}^{N_q}$ denote an arbitrary row vector extracted from $X_{unf}$. Subsequently, $x'$ is mapped into a quantum state $|\psi(x')\rangle$ in a Hilbert space. A commonly employed approach for this mapping is angle encoding, defined as:

$$|\psi(x')\rangle = \left(\bigotimes_{j=1}^{N_q} R_y(x'_j)\right)|0\rangle^{\otimes N_q} \tag{1.2}$$

where $R_y(\theta) = e^{-i\theta\sigma_y/2}$ represents a single-qubit rotation about the $y$-axis by angle $\theta$, and $\sigma_y$ denotes the Pauli-Y operator. Here, $x'_j$ is the $j$-th component of the vector $x'$. The prepared state $|\psi(x')\rangle$ is then fed into a pre-constructed quantum circuit. To capture diverse patterns analogous to multiple convolutional kernels in classical neural networks, we employ $N$ mutually independent randomized quantum circuits (in this work, we set $N = 4$). The evolution of the quantum state through the $i$-th circuit, denoted as $N_i$, is governed by a sequence of single-qubit rotation gates and two-qubit entangling gates (e.g., CNOT). The unitary operation of the $i$-th circuit can be expressed as:

$$|\phi_i\rangle = N_i |\psi(x')\rangle, \quad i = 1, 2, \ldots, N \tag{1.3}$$

The locality of the operation is preserved by restricting the entangling gates to act only on adjacent qubits. Following the circuit evolution, a quantum measurement is performed to extract features. We compute the expectation value of a measurement operator $O$ (e.g., the Pauli-Z operator on the first qubit) for each circuit:

$$y_i = \langle \phi_i | O | \phi_i \rangle \tag{1.4}$$

Finally, the scalar outputs $y_i$ from all $K$ circuits are concatenated and reshaped to form the final feature map $Y \in \mathbb{R}^{B \times K \times T' \times S'}$.

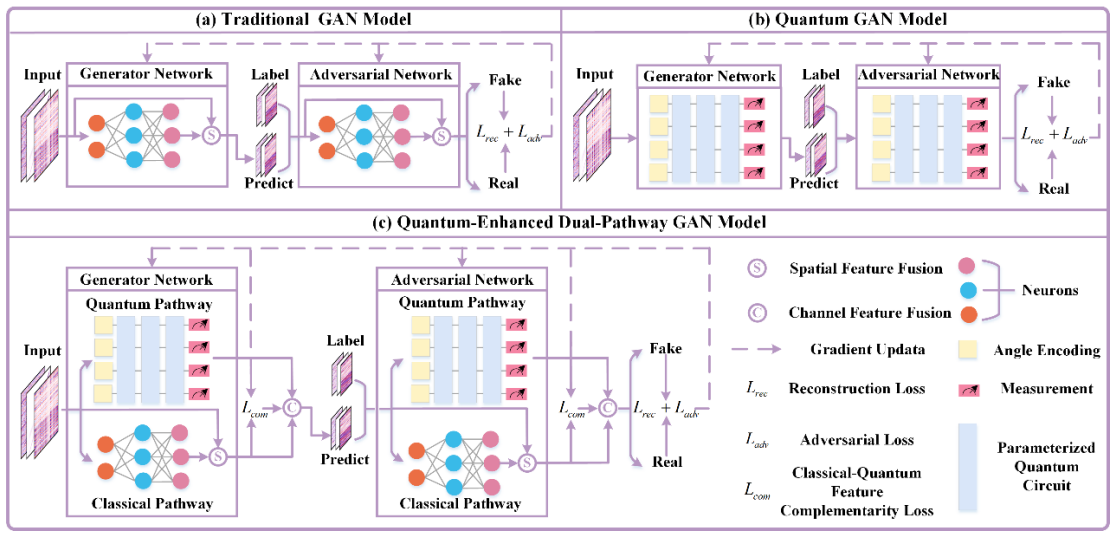

Figure 1. Comparison of GAN models relevant to our approach: (a) classical perceptron-based GAN; (b) GAN entirely implemented using quantum NNs; (c) the proposed QC-GAN.

**QC-GAN**

The primary objective of QC-GAN is to augment the representational capacity of classical NNs by embedding quantum-state features, thereby enhancing model performance in seismic data restoration tasks. To achieve this, the proposed architecture comprises two core components: the discriminator, $D_{hyb}$, and the generator, $G_{hyb}$, both of which employ a dual-pathway feature fusion strategy. As shown in Fig 1 (c). Compared with the classical GAN based on perceptrons (panel (a)), QC-GAN provides enhanced feature representation and collaborative learning capabilities. In contrast to the fully quantum GAN shown in panel (b), this hybrid design enables the efficient processing of large-scale seismic datasets without being constrained by the availability of quantum hardware, while still leveraging quantum-enhanced representational power. $D_{hyb}$ is designed to strictly distinguish between authentic, high-fidelity seismic records and the synthetic restorations produced by the generator from noise-contaminated or incomplete observations. Conversely, $G_{hyb}$ is trained to "fool" the discriminator by synthesizing outputs that preserve the structural patterns and amplitude characteristics of real seismic data. The optimization objective of QC-GAN comprises three distinct components: the generator loss ($L_G$), the adversarial loss for the discriminator ($L_{adv}$), and the convolutional-quantum feature complementarity loss ($L_{com}$). $L_G$ ensures that the generated seismic data remains consistent with the ground truth in terms of both statistical distribution and pixel-level waveform structures. It combines an adversarial term with an $L_1$ norm constraint:

$$L_G = -\mathbb{E}_{X^d}\left[\log D_{hyb}(G_{hyb}(X^d))\right] + \lambda_{rec}\mathbb{E}_{X^d,X^t}\left\|G_{hyb}(X^d) - X^t\right\|_1 \quad (1.5)$$

where $X^d \in \mathbb{R}^{B \times C \times T \times S}$ denotes the degraded observational data (e.g., noise-contaminated or missing traces), and $X^t \in \mathbb{R}^{B \times C \times T \times S}$ represents the ideal target data. The term $-\mathbb{E}_{X^d}[\log D_{hyb}(G_{hyb}(X^d))]$ represents the adversarial loss component, encouraging the generator to produce outputs classified as real by the discriminator. The term $\lambda_{rec}\|G_{hyb}(X^d) - X^t\|_1$ imposes a pixel-wise consistency constraint. $L_{adv}$ follows the standard game-theoretic strategy of GAN, aiming to maximize the probability of correctly identifying real samples while detecting generated counterparts:

$$L_{adv} = \mathbb{E}_{X^t}\left[\log D_{hyb}(X^t)\right] + \mathbb{E}_{X^d}\left[\log\left(1 - D_{hyb}\left(G_{hyb}(X^d)\right)\right)\right] \tag{1.6}$$

Here, maximizing $\mathbb{E}_{X^t}[\log D_{hyb}(X^t)]$ ensures the discriminator accurately captures the distribution of genuine seismic signals, while the second term strengthens its ability to detect synthetic data. To prevent feature redundancy between the quantum and classical pathways, we introduce the QC feature complementarity loss, $L_{com}$. This loss enforces orthogonality between the two feature sets, compelling the quantum pathway to capture high-dimensional patterns that are challenging for classical NNs. It is formulated as the absolute cosine similarity between the classical features $X$ and quantum features $\mathcal{Q}$:

$$L_{com} = |\cos(X, \mathcal{Q})| = \frac{|X \cdot \mathcal{Q}|}{\|X\| \|\mathcal{Q}\|} \tag{1.7}$$

Minimizing $L_{com}$ during training mitigates the correlation between the dual pathways, ensuring the extraction of complementary features.

The overall operational workflow of the proposed QC-GAN is illustrated in Fig 2. Accordingly, the generative process of $G_{hyb}$ is formally defined as follows:

$$X^p = G_{hyb}(X^d) = f^{up}\left(X^0 + \mathcal{F}_{res}(X^0)\right) \tag{1.8}$$

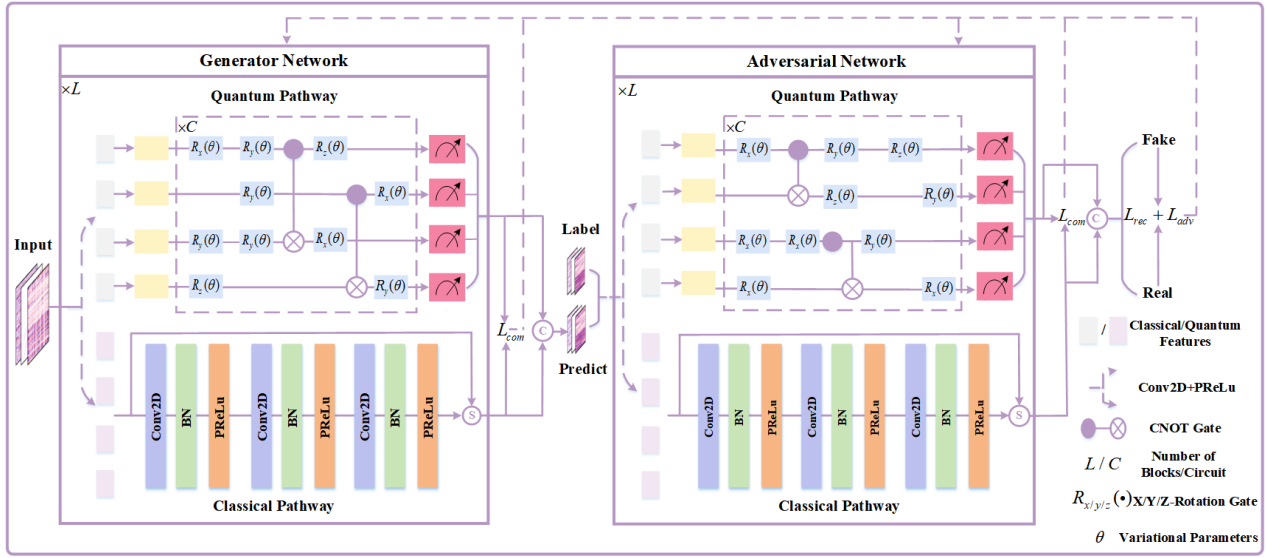

Figure 2. Schematic illustration of the proposed QC-GAN architecture. The framework integrates a QC generator and discriminator, both employing dual-pathway feature fusion strategies for seismic data processing.

Initially, the degraded input $X^d$ is processed by a preliminary convolutional layer followed by a Parametric Rectified Linear Unit (PReLU) to yield the initial feature map $X^0$. Subsequently, $X^0$ is fed into the $\mathcal{F}_{res}$ block, which comprises parallel quantum and classical pathways:

$$\mathcal{F}_{res}(X^0) = \sum_{l=1}^{L} \Re^l\left(X^{l-1}\right) \oplus Q^{l-1} \tag{1.9}$$

In the $l$-th block ($l = 1, \ldots, L$), the input $X^{l-1}$ is split along the channel dimension into two partitions, which are subsequently fed into parallel classical ($\Re^l$) and quantum pathways, respectively. Prior to fusion, the extracted classical features and the generated quantum features are utilized to compute the QC feature complementarity loss $L_{com}$ (Eq. 7) to enforce feature orthogonality. Subsequently, these two feature sets are integrated via channel-wise concatenation ($\oplus$). The classical pathway $\mathfrak{R}^l$ is designed with a deep residual architecture to enhance feature extraction capabilities. Specifically, it employs a cascade of three convolutional units, formulated as:

$$\mathfrak{R}^l\left(X^{l-1}\right) = X^{l-1} + \Phi_3\left(\Phi_2\left(\Phi_1(X^{l-1})\right)\right) \tag{1.10}$$

Each $\Phi_i(\cdot)$ represents a composite function comprising a convolutional layer, Batch Normalization (BN), and PReLU activation. This stacked architecture facilitates the capture of complex non-linear seismic patterns prior to residual aggregation. The final aggregated features are then processed by the up-sampling module $f^{up}$, which utilizes pixel shuffle and convolution operations to map the latent features to high-fidelity restorations. This hybrid architecture effectively preserves the Spatiotemporal feature extraction advantages of classical NNs while leveraging the high-dimensional mapping capabilities of quantum NNs.

The discriminator, denoted as $D_{hyb}$, adopts a dual-path structural philosophy analogous to that of $G_{hyb}$, incorporating a cascade of $L$ $\mathcal{F}_{res}$ blocks to ensure stable gradient propagation and effective feature extraction. The input data (whether a ground truth $X^t$ or a generated sample $X^p$) is initially processed to yield the base feature map $X^0$. Mirroring the generator's architecture, this feature map is simultaneously propagated through parallel classical and quantum pathways, where classical features are integrated with the corresponding quantum features along the channel dimension. However, a key distinction lies in the final stage: the output of the last $\mathcal{F}_{res}$ block, denoted as $X^L \oplus Q^L$, is flattened and processed by a fully connected layer ($f_{fc}$) followed by a Sigmoid activation to compute the probability of authenticity:

$$D_{hyb}(X^p) = \sigma\left(f_{fc}\left(X^L \oplus Q^L\right)\right) \tag{1.11}$$

This residual-based decision-making mechanism significantly enhances the discriminator's capability to discern complex seismic structural patterns while effectively mitigating the vanishing gradient problem prevalent in deep adversarial networks.

## EXPERIMENTS

In this section, we conduct a series of comparative experiments on field seismic data to evaluate the interpolation and denoising performance of QC-GAN. Additionally, we also combine the proposed dual-path QC synergistic framework with the classical UNet and apply the QC-UNet to the task of low-frequency extrapolation to validate the architectural versatility and task-oriented generalizability. All experiments were implemented using the PyTorch framework on an Intel Xeon Gold 6348 CPU (2.60 GHz) paired with an NVIDIA RTX A800 GPU for acceleration. The quantum NN components realized via the DeepQuantum library (the source code link is provided in the Data and Materials Availability section). For network training, we employed the Adam optimizer. The training procedure spanned 100 epochs with a batch size of 256. Specifically, the learning rates were set to $1\times10^{-5}$ for QC-GAN and $1\times10^{-4}$ for QC-UNet. To ensure a rigorous and fair comparison, all baseline and proposed methods were trained under identical settings regarding iteration counts, sample partitioning, and experimental protocols.

**Seismic data interpolation**

As a critical pre-processing step, Seismic data interpolation aims at recovering missing information from sparse or incomplete observational data. In this subsection, we utilize the Parihaka dataset with a temporal sampling interval of 4 ms, to test the effectiveness of QC-GAN in seismic data interpolation. Specifically, sliding window cropping operations are applied to the dataset to generate 100000 patches with a size of 224×128 (time × trace). These patches are then partitioned into training, validation, and testing sets at a ratio of 8:1:1. To simulate sparse observational geometries, the training and validation sets are constructed under a random missing condition, where 30% to 70% of the traces are randomly removed from each patch. During the training phase, Mean Absolute Error (MAE) and Root Mean Square Error (RMSE) are employed to quantitatively monitor the interpolation fidelity. Detailed mathematical definitions of these metrics are provided in the discussion section. As depicted in Fig. 3, (a) and (b) correspond to the error evolution on the training set, whereas (c) and (d) panel illustrate the performance on the validation set. Within each subplot, the learning curves of the classical GAN and QC-GAN are compared. A marked quantitative improvement is evident when transitioning from the purely classical paradigm to the QC-synergistic architecture. Relative to the classical GAN, the QC-GAN achieves a substantially lower error floor, lowering the MAE from approximately 0.04 to 0.010 and the RMSE from 0.45 to 0.02 on both datasets. This reduction in error magnitudes demonstrates that the embedded quantum-state features effectively enhance the precision of seismic data interpolation.

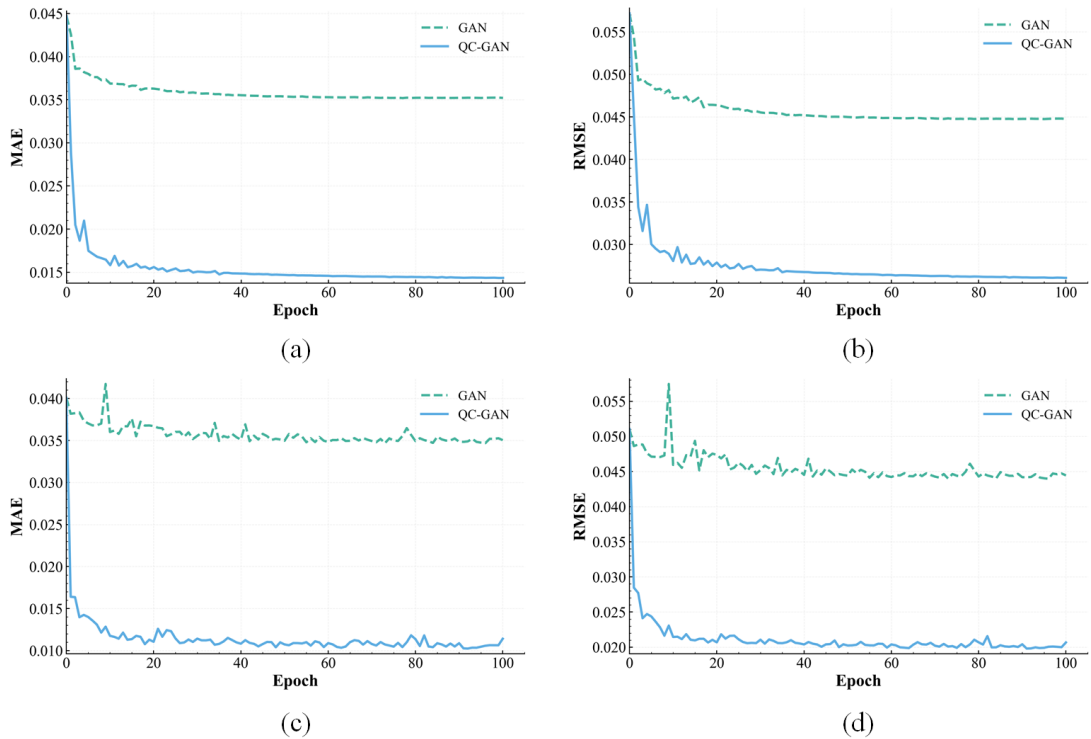

Figure 3. Comparison of loss curves between the classical GAN and the proposed QC-GAN: (a–b) training, (c–d) validation.

To comprehensively evaluate the performance of QC-GAN, the testing set incorporates two distinct missing patterns: regular missing and random missing, which are distributed in an equal ratio within the test set. A randomly selected complete test patch is displayed in Fig. 4(a) as the ground truth. For the regular missing condition, we apply a regular trace mask (removing one trace after every two intact traces), yielding the incomplete patch shown in Fig. 4(b). For the random missing condition, we adopt the same missing pattern as that used during the training and validation stages, where 30%–70% of the traces are randomly removed from each patch, resulting in the sparse patch illustrated in Fig. 4(c).

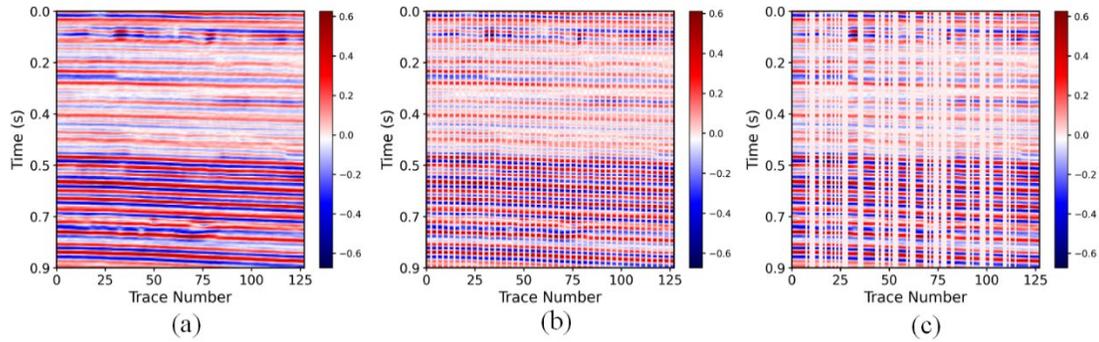

Figure 4. Illustration of the missing patterns. (a) Complete test patch. (b) Incomplete test patch generated under the regular missing condition. (c) Incomplete test patch generated under the random missing condition.

The interpolation results are presented in Fig. 5. As shown in (a) and (b), both methods are able to recover the incomplete traces to some extent. However, from the region indicated by the red arrows, we can observe that QC-GAN exhibits substantially lower residual energy than the classical GAN, demonstrating the improved interpolation fidelity. This is highlighted in the corresponding residual maps ((c) and (d) panels), which reveal discrepancies in preserving the seismic reflection amplitudes. As indicated by the red arrows, QC-GAN exhibits substantially lower residual energy than the classical GAN, demonstrating the improved interpolation fidelity. (e) and (f) present the interpolation results under the random missing pattern, with (g) and (h) depicting the associated residual maps. As indicated by the black arrows, the classical GAN demonstrates distinct limitations in amplitude preservation, suffering from severe coherent energy leakage. Specifically, a substantial amount of coherent seismic events fails to be effectively reconstructed, leaking directly into the residual profiles. This phenomenon suggests that the classical architecture struggles to fully capture the complex structural morphologies and amplitude variations of the seismic data. In stark contrast, the QC-GAN demonstrates enhanced amplitude preservation, yielding the minimum residual magnitude and effectively restoring the original intensity and continuity of the reflections, even under severely irregular sampling conditions.

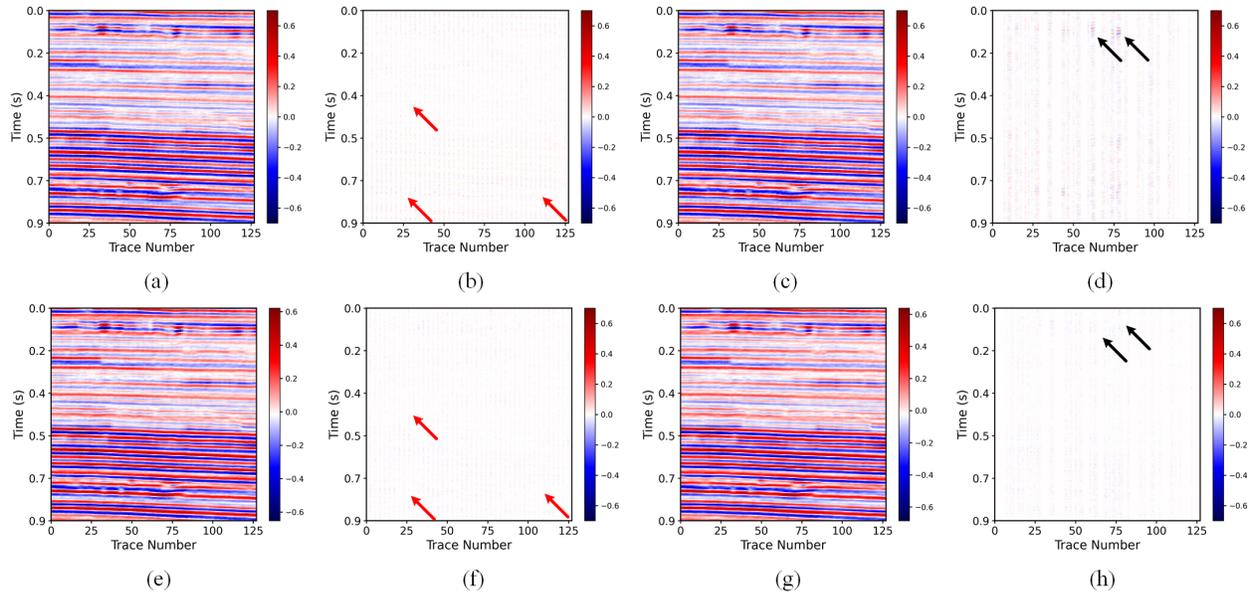

Fig 5. (a)-(d) and (e)-(h) illustrate the interpolation results and corresponding residual maps under regular and random missing patterns using the classical GAN and the QC-GAN, respectively.

To further evaluate the interpolation fidelity, we extracted individual traces from the interpolated results for a detailed waveform comparison, as illustrated in Figure 6. Under the regular missing condition, (a) and (b) display the waveform alignment between the ground truth and the single-trace interpolation results from the classical GAN and QC-GAN, respectively. To quantify the deviation between these interpolated results and the ground truth, we plotted their absolute residuals in (e). Correspondingly, under the random missing condition, (c) and (d) illustrate the waveform matching for the classical GAN and QC-GAN, while (f) presents the associated absolute residuals. Consistent with the observations in Figure 5, the traces interpolated by QC-GAN (red dash-dotted line) exhibit a higher agreement with the ground truth (black solid line) compared to those from the classical GAN (blue dashed line). QC-GAN accurately reproduces the amplitude peaks and waveform phases without noticeable distortion. In contrast, the classical GAN displays notable amplitude underestimation at prominent seismic events, further corroborating the signal leakage issue identified in the residual maps of Figure 5.

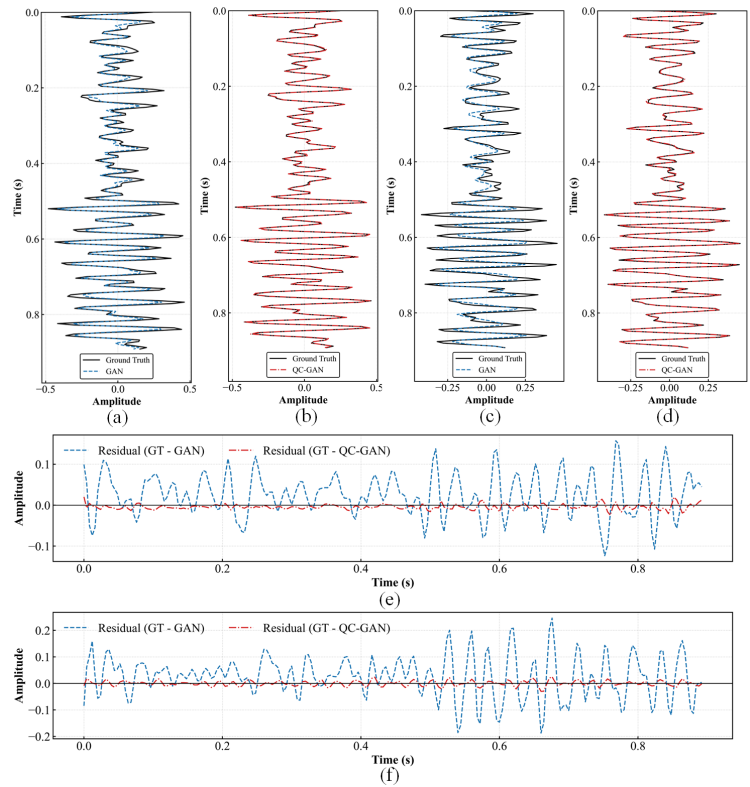

Figure 6. Trace comparisons and corresponding residuals of the classical GAN and QC-GAN relative to the ground truth under regular [(a)-(b), (e)] and random [(c)-(d), (f)] missing conditions.

We further conducted spectrum comparisons as illustrated in Figure 7. Specifically, (a) and (b) present the amplitude spectra of the interpolated traces versus the ground truth for both the classical GAN and QC-GAN under the regular missing condition, while (c) and (d) display the corresponding comparisons under the random missing condition. As observed from the spectra, the QC-GAN results (red dash-dotted line) exhibit a high degree of overlap with the ground truth across the entire frequency bandwidth, particularly in preserving high-frequency components. In contrast, the classical GAN shows a noticeable energy decay in the mid-to-high frequency range in (a). Moreover, significant spectral artifacts are observed at certain frequency bands where the interpolated amplitude exceeds the ground truth in (c), suggesting that the classical GAN introduces non-physical spurious energy. This further demonstrates that the QC-GAN is capable of effectively maintaining both structural and spectral integrity.

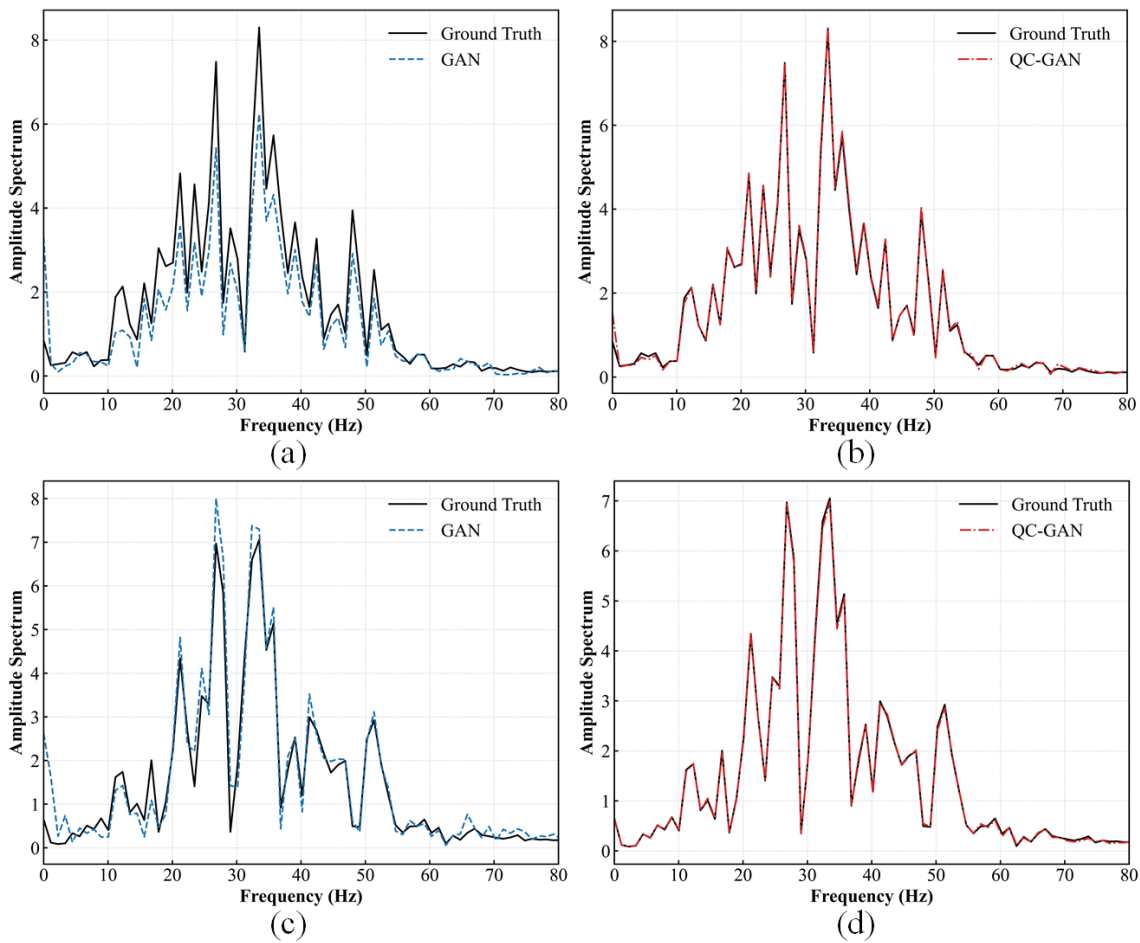

Figure 7 spectrum comparisons between the interpolated trace and the ground truth. (a) - (b) show the classical GAN and QC-GAN, respectively, under the regular missing condition; (c) - (d) show the classical GAN and QC-GAN, respectively, under the random missing condition.

To systematically corroborate the aforementioned visual observations, we conduct a quantitative numerical analysis by comparing the interpolated results against the ground truth across four standard numerical metrics: MAE, RMSE, Peak Signal-to-Noise Ratio (PSNR), and Structural Similarity Index Measure (SSIM). Detailed mathematical definitions of PSNR and SSIM, as well as MAE and RMSE, are provided in the Discussion section. Table 1 presents a quantitative comparison of the interpolation performance under the regular and random missing trace patterns for the test patches shown in Fig. 5. QC-GAN consistently achieves lower MAE and RMSE values than the classical GAN regardless of the missing pattern. Quantitatively, the QC-GAN achieves substantially higher PSNR and SSIM scores than the classical GAN. These elevated metrics demonstrate its superior capability in preserving the spatial continuity and amplitude fidelity of coherent seismic events. This improvement is attributed to the synergistic integration of high-dimensional quantum-state representations, which effectively mitigates coherent energy leakage even under severely degraded acquisition conditions.

Table 1: Quantitative comparison of interpolation performance under regular and random missing patterns on the test patches.

|         | Method  | MAE    | RMSE   | PSNR    | SSIM   |
|---------|---------|--------|--------|---------|--------|
| Regular | GAN     | 0.0078 | 0.0101 | 42.0782 | 0.9771 |
|         | QC-GAN  | 0.0060 | 0.0077 | 44.4564 | 0.9840 |
| Random  | GAN     | 0.0109 | 0.0147 | 38.3814 | 0.9411 |
|         | QC-GAN  | 0.0096 | 0.0129 | 39.4488 | 0.9503 |

**Seismic data denoising**

Seismic data denoising is a fundamental prerequisite for reliable geological interpretation, aiming to suppress noise while preserving valid signal amplitudes and structural continuity. To assess the QC-GAN under low signal-to-noise ratio conditions, we constructed a dataset derived from the Kerry dataset. Specifically, 6,9000 seismic patches, each with dimensions of 224×128 (time × traces), were randomly extracted from the Kerry dataset. To generate the noisy dataset for supervised learning, additive white Gaussian noise with a zero mean and a standard deviation of 0.1 was injected into these clean samples. The dataset was subsequently partitioned into training, validation, and test sets following an 8:1:1. Fig 8 (a) presents the ground truth of a representative test patch, while Fig 8 (b) shows its noise-contaminated counterpart, which corresponds to a low SNR of 0.69 dB.

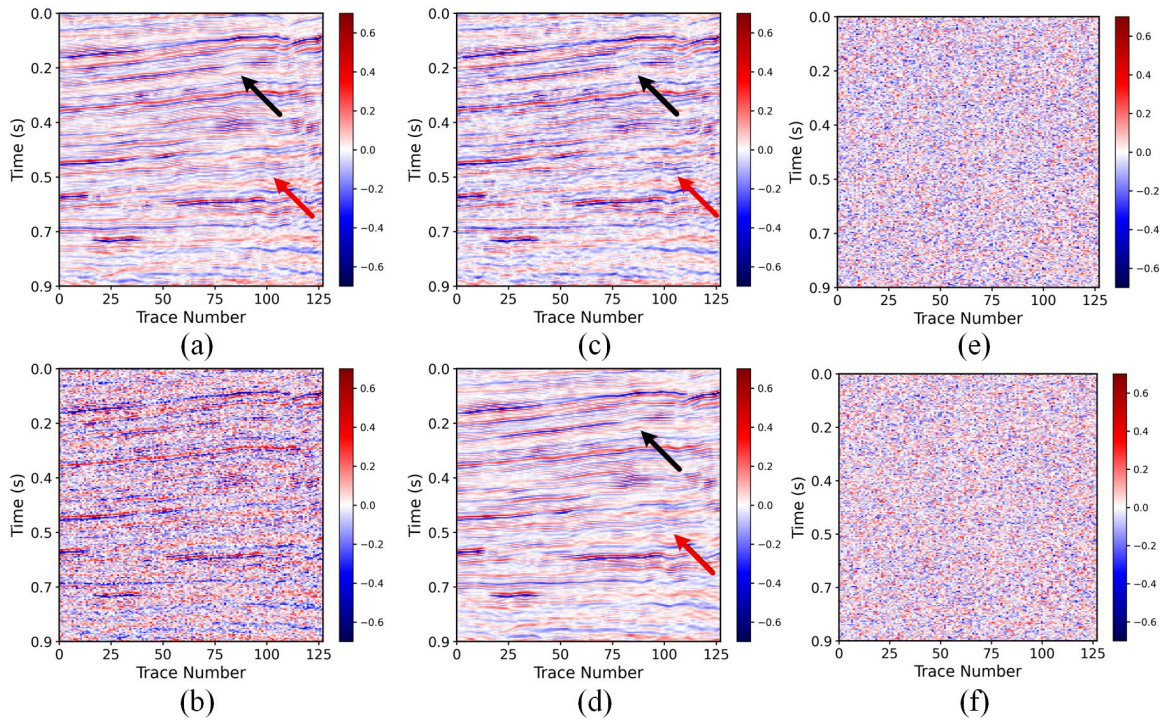

Figure 8. (a) Clean seismic patch. (b) Noisy input. (c) Classical GAN denoised result. (d) QC-GAN denoised result. (e) and (f) display the absolute residual maps of (c) and (d) relative to the noisy input in (b), respectively.

Evidently, the injection of random noise heavily obscures crucial seismic reflection events, particularly causing significant detail loss in deeper weak-signal regions. Figs. 8 (c) and (d) depict the denoising results obtained from the classical GAN and the QC-GAN, respectively. Furthermore, (e) and (f) display the absolute residual maps of (c) and (d) relative to the noisy input in (b). As observed from the residual maps, neither method exhibits severe coherent signal leakage. However, a closer inspection reveals that the classical GAN in (c) fails to preserve the spatial continuity of the seismic events in multiple regions compared to QC-GAN in (d). This deficiency leads to noticeable structural fragmentation, as highlighted by the black and red arrows, coupled with pervasive spatial blurring across the entire profile. In stark contrast, the QC-GAN effectively mitigates this phenomenon. By leveraging high-dimensional quantum features to encode the non-linear dynamics of weak reflections, the proposed architecture recovers deep seismic structures with superior clarity and strictly preserves their spatial continuity.

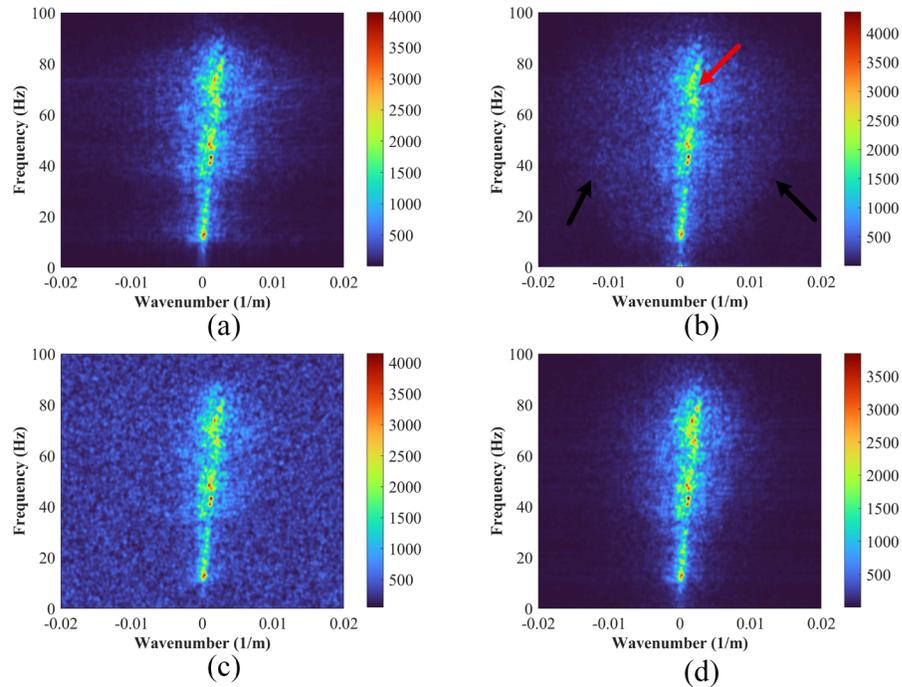

Figure 9. Frequency spectra of (a) the clean seismic patch, (b) the noisy input, (c) the classical GAN denoised result, (d) the QC-GAN denoised result.

Furthermore, to evaluate the signal fidelity from a frequency perspective, we conducted a spectral analysis on the data from Figs. 8(a)–(d), with the results presented in Fig. 9. Specifically, upon the addition of random noise (Fig. 9(b)), the original frequency spectrum (Fig. 9(a)) is severely distorted, with the effective signals heavily masked by noise. Following the denoising process, the classical GAN suffers from a noticeable loss of high-frequency energy (indicated by the red arrow in Fig. 9(c)). Additionally, it fails to completely suppress the interference, leaving behind prominent residual noise artifacts (highlighted by the black arrow). This frequency-domain deficiency directly corroborates the spatial blurring and structural fragmentation observed earlier. In contrast, as depicted in Fig. 9(d), the QC-GAN successfully yields an amplitude spectrum that closely aligns with that of the ground truth. To evaluate signal fidelity at the trace level, a randomly selected seismic trace was extracted for waveform analysis, as shown in Fig. 10. The waveform comparison demonstrates that the QC-GAN recovery achieves precise phase alignment and high amplitude fidelity with respect to the ground truth, thereby confirming the robustness of the QC-GAN in preserving subtle waveform characteristics.

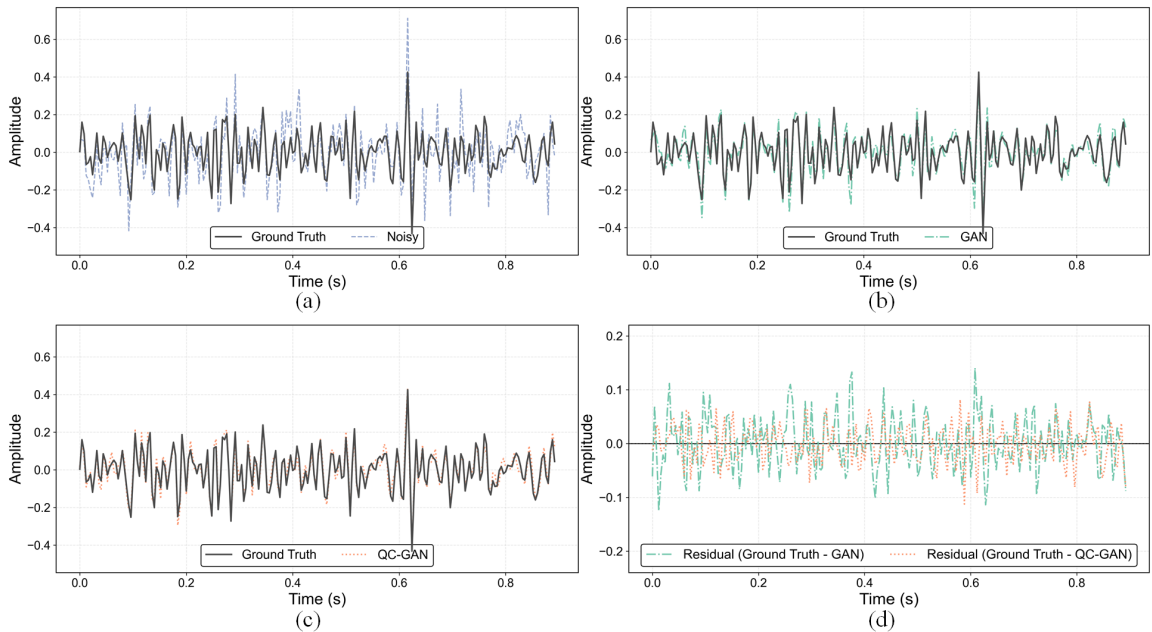

Figure 10. Trace comparisons relative to the ground truth: (a) noisy input, (b) classical GAN, and (c) QC-GAN, with (d) the corresponding denoising residuals.

**Low-frequency extrapolation**

Field seismic data frequently suffer from a deficiency in low-frequency information due to instrumental bandwidth limitations or ambient noise. This absence often causes full waveform inversion to stagnate in local minima due to the cycle-skipping phenomenon. Consequently, low-frequency extrapolation (LFE) is employed to recover the missing spectral bands. To validate the effectiveness and generalizability of our proposed QC synergistic mechanism across different architectures and tasks, we extended the framework to LFE. Specifically, we transitioned from a GAN-based backbone to a classical UNet architecture. The UNet's encoder-decoder structure serves as the classical convolutional pathway, which is then integrated with the quantum pathway to construct the QC-UNet.

To construct the training and validation datasets for QC-UNet, we initiated the process by partitioning the Marmousi velocity model into 16 patches using a $32 \times 96$ rectangular sliding window. Four adjacent sub-models were combined and interpolated to form complex sub-models. Seismic data were then generated using a finite-difference scheme based on the acoustic wave equation with perfectly matched layer boundary conditions. Modeling parameters included a grid spacing of 20 m, a sampling interval of 0.002 s, and a recording duration of 4.8 s, yielding a total of 921,600 trace pairs (split 9:1 for training, validation). Crucially, to explicitly formulate the LFE task, the supervised learning pairs were designed with distinct spectral characteristics: the input data were generated using a 7 Hz Ricker wavelet followed by a high-pass filter to remove information below 5 Hz, thereby simulating band-limited observations lacking low-frequency content. Conversely, the label data utilized the same wavelet but were subjected to a 5 Hz low-pass filter. This configuration compels the network to learn the non-linear mapping required to extrapolate the target low-frequency background solely from the available high-frequency features.

To evaluate the generalization of QC-UNet in practical applications, field seismic data from an OBC survey were employed as the test dataset. the pressure (P) component of the four-component OBC data was extracted, and the data preprocessing workflow remained consistent with the previous sections. The data features a sampling interval of 2 ms and a receiver interval of 25 m. Fig. 11(a) displays the input data with a frequency band of 5–10 Hz obtained via band-pass filtering, while Fig. 11(b) shows the ground-truth low-frequency data extracted by applying a 0–5 Hz low-pass filter to the original broadband data. Fig. 11(c) presents the prediction result of the classical U-Net. It is evident that this method fails to effectively recover the coherent low-frequency signals in several regions (indicated by black arrows) and introduces pronounced artifacts (highlighted by the red rectangular boxes). In contrast, the predicted result of the QC-UNet (Fig. 11(d)) not only yields clear and continuous seismic events but also demonstrates superior recovery accuracy and amplitude preservation, particularly in the large-offset and late-arrival regions.

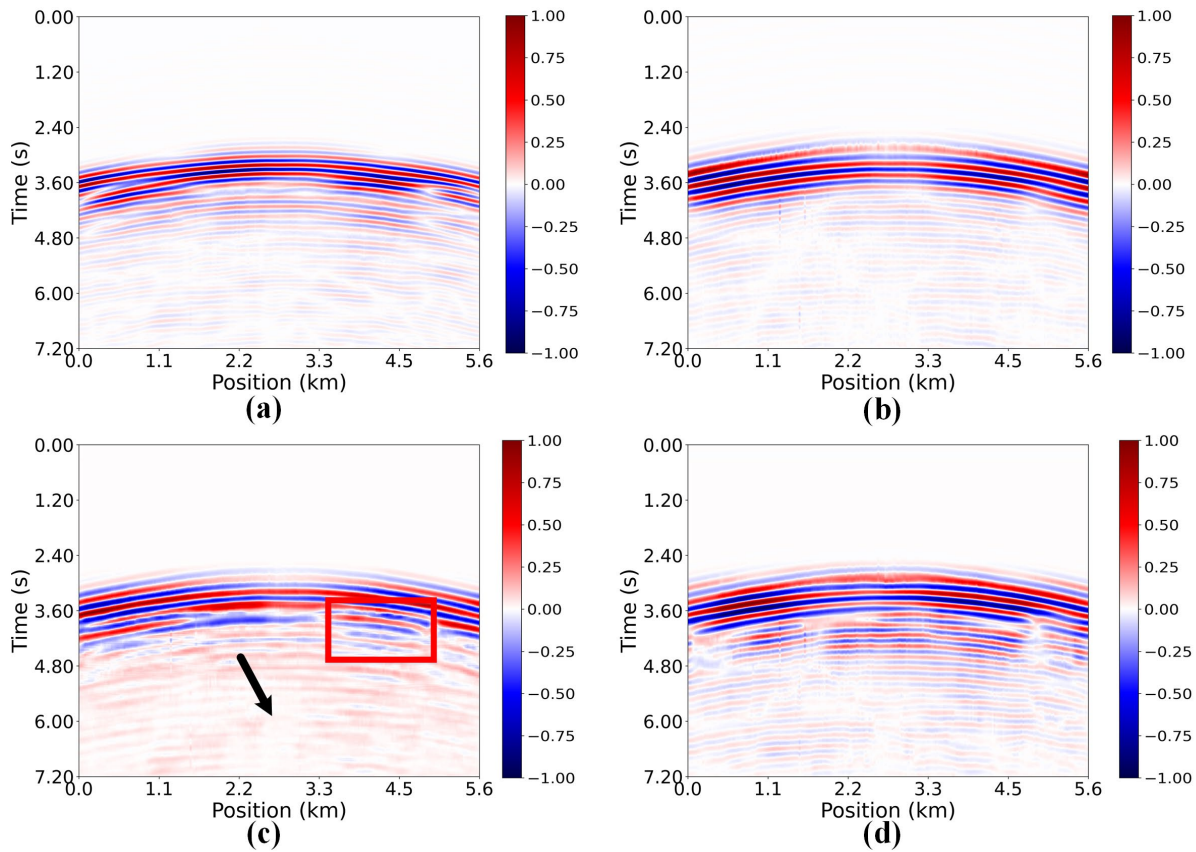

Figure 11. (a) Input data (5‐10Hz). (b) Ground truth (0–5 Hz). (c) Classical UNet LFE result. (d) QC-UNet LFE result.

Furthermore, the frequency-wavenumber spectra in Fig. 12 were analyzed to evaluate the spectral consistency of the results. (a) and (b) represent the F-K spectra of the input data and the ground-truth low-frequency data, respectively. (c) shows the F-K spectrum predicted by the classical U-Net; although it manages to extrapolate a portion of the low-frequency energy, it simultaneously generates distinct artifacts within the low-frequency band (as indicated by the red arrow). Conversely, the F-K spectrum from the QC-UNet prediction (Fig. 12(d)) not only effectively suppresses these artifacts but also exhibits a clearer distribution of low-frequency energy.

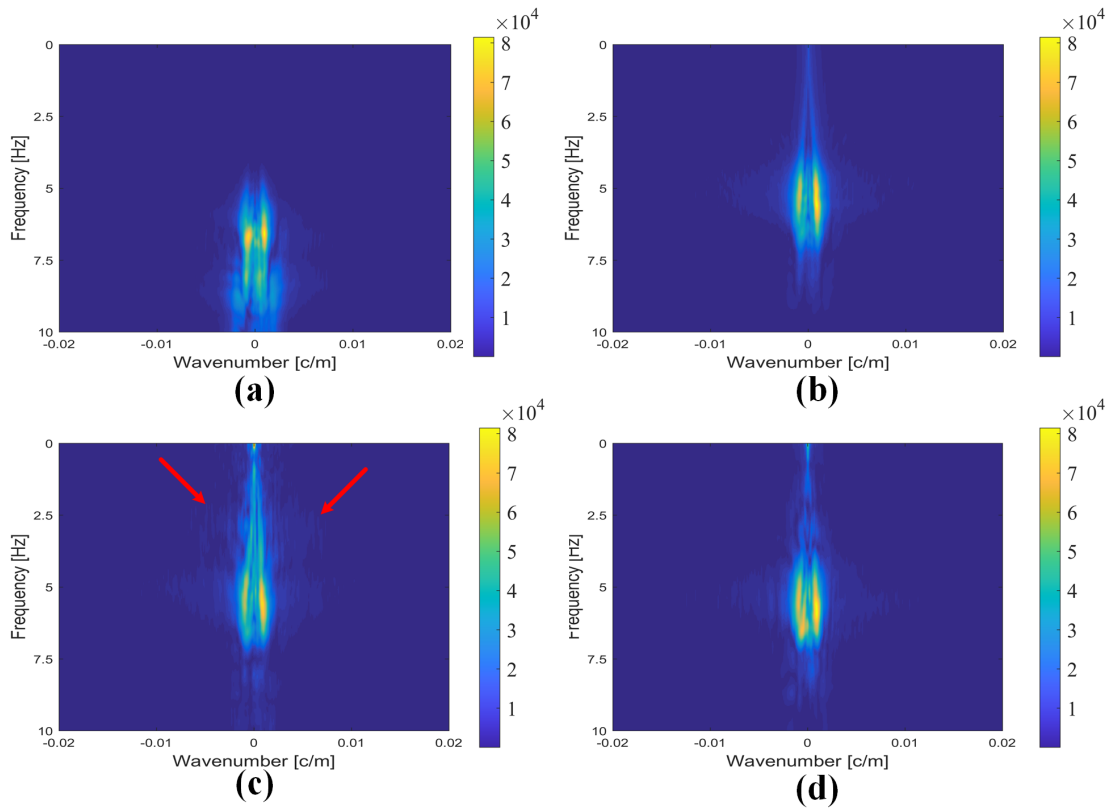

Figure 12. Frequency spectra of (a) the input data, (b) the ground truth, (c) the classical UNet LFE result, and (d) the QC-UNet LFE result.

To quantitatively evaluate the amplitude fidelity from a trace-level perspective, we extracted individual traces from the predicted results of both the classical U-Net and the QC-UNet. The single-trace residuals were then calculated by computing the absolute differences between these predictions and the corresponding ground truth. As illustrated in Fig. 13, (a) displays the single-trace comparison between the input data and the ground truth, while (b) and (c) compare the predictions of the classical U-Net and QC-UNet against the ground truth, respectively. (d) presents a comparison of the residuals for both methods. As observed from these plots, the classical U-Net exhibits significant amplitude discrepancies. In contrast, the residuals of the QC-UNet remain at a consistently minimal level, demonstrating its improved amplitude preservation and waveform fidelity.

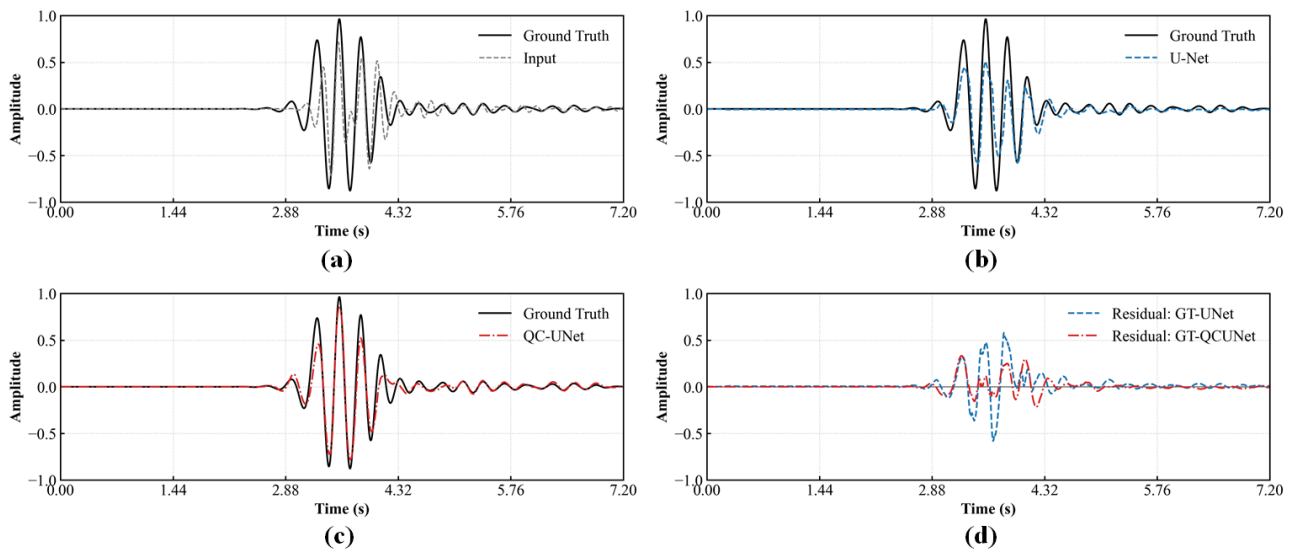

Figure 13. Trace comparisons relative to the ground truth: (a) input, (b) classical U-Net, (c) QC-UNet, (d) the corresponding residuals.

Furthermore, to verify the accuracy of the recovered frequency components, we conducted a spectrum comparison for these extracted traces, as shown in Fig. 14. Specifically, (a) illustrates the spectral gap between the input data and the ground truth, highlighting the missing low-frequency information. (b) and (c) display the predicted spectra of the classical U-Net and the QC-UNet compared with the ground-truth spectrum, respectively. It is evident that the QC-UNet spectrum aligns more closely with the ground truth across the target 0–5 Hz band, whereas the classical U-Net fails to fully recover the low-frequency energy and introduces spectral leakage.

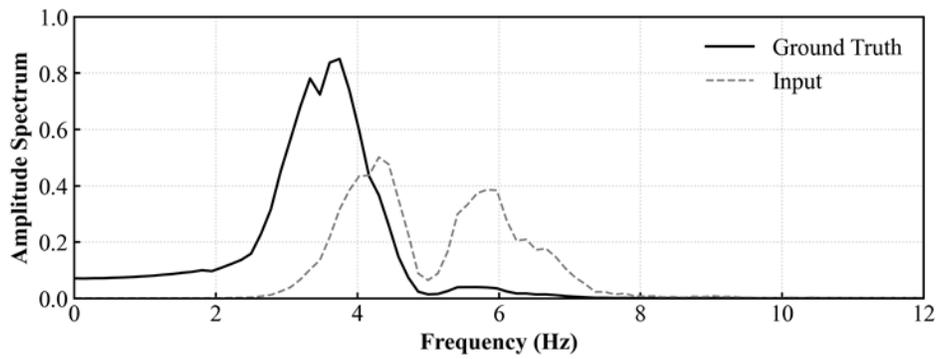

(a)

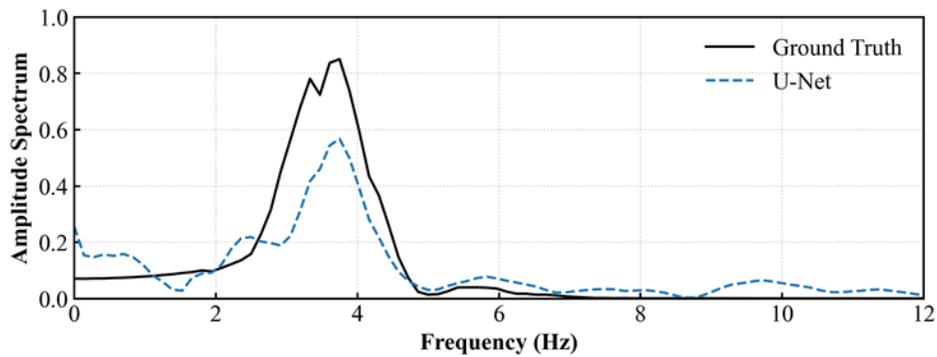

(b)

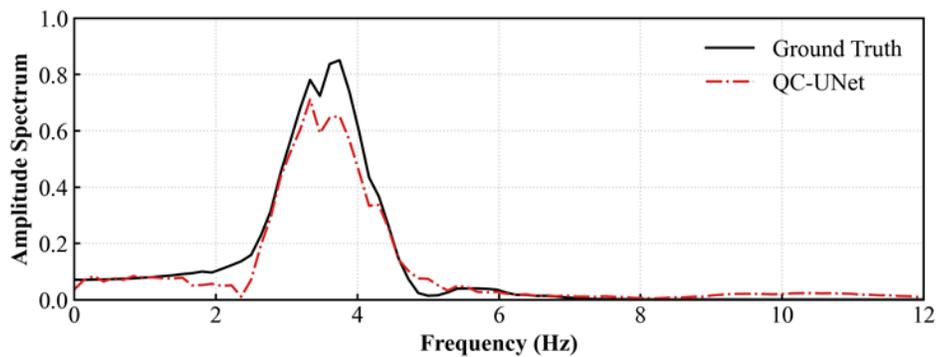

(c)

Geophysics                                                                                                         30

Figure 14. Spectrum comparisons relative to the ground truth: (a) input, (b) classical U-Net, (c) QC-UNet.

## CONCLUSION

This study proposes a QC synergistic paradigm to address the inherent representational limitations of classical NNs in seismic data processing. By integrating quantum NNs within a GAN, the proposed QC-GAN leverages the high-dimensional mapping capabilities of the Hilbert space to capture complex seismic features that are elusive to classical NNs. To optimize the fusion of quantum and classical features, we introduce a QC feature complementarity loss grounded in orthogonalization. This constraint explicitly enforces the QC-GAN to learn non-redundant, synergistic representations. Experiments on field data demonstrate that this paradigm maintains superior amplitude fidelity in both seismic data interpolation and denoising tasks. The QC synergistic paradigm's versatility is further substantiated by its extension to a U-Net architecture; this QC-U-Net demonstrates substantial improvements over classical NNs in LFE. Collectively, our findings highlight the potential of this synergistic approach as a robust and scalable paradigm for next-generation seismic data processing.

## DISCUSSION

Here, we detail the evaluation metrics used to quantify the performance of our models in the experiments section, including MAE, RMSE, PSNR, and SSIM. Let $y_i$ denote the ground-truth data, $\hat{y}_i$ denote the reconstructed data, and $N$ represent the total number of sampling points. MAE measures the average magnitude of absolute errors between the reconstructed data and the ground truth, directly reflecting the actual deviation of the model's output. A lower MAE indicates higher reconstruction accuracy. The formula is:

$$MAE = \frac{1}{N}\sum_{i=1}^{N}|y_i - \hat{y}_i| \qquad (1.12)$$

RMSE calculates the square root of the average squared differences between the predicted and actual values. Compared to MAE, RMSE applies a heavier penalty to larger errors, making it more sensitive to local anomalous deviations in the reconstructed results. It is widely used to assess the global stability of the reconstruction process. The formula is:

$$RMSE = \sqrt{\frac{1}{N}\sum_{i=1}^{N}(y_i - \hat{y}_i)^2} \qquad (1.13)$$

PSNR is a standard objective criterion for evaluating signal reconstruction quality. It defines the ratio between the maximum possible power of a signal and the power of corrupting noise that affects its fidelity. Measured in decibels (dB), a higher PSNR value indicates lower distortion and suppressed noise in the reconstructed data. $MAX_Y$ represents the maximum possible amplitude of the data. The formula is:

$$PSNR = 10\log\left(\frac{MAX_Y}{RMSE}\right) \qquad (1.14)$$

SSIM comprehensively evaluates the spatial feature similarity between two sets of data across three dimensions: luminance (mean), contrast (variance), and structure (covariance). The SSIM value ranges from [0, 1], where a value closer to 1 implies a higher structural consistency between the reconstructed results and the ground truth. The formula is:

$$SSIM(y, \hat{y}) = \frac{(2\mu_y \mu_{\hat{y}} + c_1)(2\sigma_{y\hat{y}} + c_2)}{(\mu_y^2 + \mu_{\hat{y}}^2 + c_1)(\sigma_y^2 + \sigma_{\hat{y}}^2 + c_2)} \tag{1.15}$$

where $\mu_y$ and $\mu_{\hat{y}}$ are the mean values of $y$ and $\hat{y}$, respectively; $\sigma_y^2$ and $\sigma_{\hat{y}}^2$ are their variances; $\sigma_{y\hat{y}}$ is the covariance; and $c_1$ and $c_2$ are small positive constants introduced to prevent the denominator from becoming zero.

## ACKNOWLEDGMENTS

This study was jointly supported by the National Natural Science Foundation of China (Grant No. 42574169), the Deep Earth Probe and Mineral Resources Exploration—National Science and Technology Major Project (Grant No. 2025ZD1007600), and the Jilin Provincial Natural Science Foundation - Free Exploration Project (Grant No. YDZJ202501ZYTS530).

## DATA AND MATERIALS AVAILABILITY

The quantum NNs components in this study were implemented using the DeepQuantum library (https://deepquantum.turingq.com). However, the complete algorithmic framework is currently undergoing intellectual property protection. Consequently, the full source code and proprietary data are not publicly available at this stage. Researchers interested in technical discussions or further clarification are encouraged to contact the corresponding author.